\documentclass{article}

\title{Disentangling feature and lazy training in deep neural networks}

\usepackage{times}
\usepackage{nicefrac}
\usepackage{pgf}
\usepackage{wrapfig}
\usepackage{import}
\usepackage[round]{natbib}
\usepackage{amsmath}
\usepackage{amssymb}
\usepackage{hyperref}
\usepackage[a4paper, total={6in, 8in}]{geometry}
\usepackage[normalem]{ulem}

\DeclareUnicodeCharacter{2212}{$-$}

\newcommand{\be}{\begin{equation}}
\newcommand{\ee}{\end{equation}}
\newcommand{\ba}{\begin{eqnarray}}
\newcommand{\ea}{\end{eqnarray}}
\newcommand{\mm}[1]{\textcolor{black}{#1}}
\newcommand{\st}[1]{\textcolor{black}{#1}}
\renewcommand{\cite}[1]{\citep{#1}}

\newcommand{\mc}{\mathcal}

\newcommand{\mr}{\mathrm}

\newcommand{\abs}[1]{|\!|#1|\!|}
\newcommand{\eref}[1]{Eq.~(\ref{#1})}

\newcommand{\taref}[1]{Table~\ref{#1}}

\newcommand{\aref}[1]{Appendix~\ref{#1}}
\newcommand{\sref}[1]{Section~\ref{#1}}
\newcommand{\fref}[1]{Fig.~\ref{#1}}

\author{%
Mario Geiger, Stefano Spigler, Arthur Jacot and Matthieu Wyart
}

\begin{document}
\maketitle

\begin{abstract}
Two distinct limits for deep learning have been derived as the network width $h\rightarrow \infty$, depending on how the weights of the last layer scale with $h$. In the Neural Tangent Kernel (NTK) limit, the dynamics becomes linear in the weights and is described by a \emph{frozen} kernel $\Theta$ (the NTK). By contrast, in the Mean-Field limit, the dynamics can be expressed in terms of the distribution of the parameters associated with a neuron, that follows a partial differential equation.
In this work we consider deep networks where the weights in the last layer scale as $\alpha h^{-\nicefrac12}$ at initialization. By varying $\alpha$ and $h$, we probe the crossover between the two limits.
We observe two the \mm{previously identified} regimes of ``lazy training'' and ``feature training''. In the lazy-training regime, the dynamics is almost linear and the NTK \mm{barely changes} after initialization. The feature-training regime includes the mean-field formulation as a limiting case and is characterized by a kernel that evolves in time, and thus learns some features.
We perform numerical experiments on MNIST, Fashion-MNIST, EMNIST and CIFAR10 and consider various architectures.
We find that: (i) The two regimes are separated by an $\alpha^*$ that scales as $\frac{1}{\sqrt{h}}$. 
(ii) Network architecture and data structure play an important role in determining which regime is better: in our tests, fully-connected networks  perform generally better in the lazy-training regime, unlike convolutional networks. 
(iii) In both regimes, the fluctuations $\delta F$ induced on the learned function by initial conditions decay as $\delta F \sim 1/\sqrt{h}$, leading to a performance that increases with $h$.
The same improvement can also be obtained at an intermediate width by ensemble-averaging several networks that are trained independently. 
(iv) In the feature-training regime we identify a time scale $t_1 \sim \sqrt{h}\alpha$, such that for $t \ll t_1$ the dynamics is linear. At $t \sim t_1$, the output has grown by a magnitude $\sqrt{h}$ and the changes of the tangent kernel $\abs{\Delta \Theta}$ become significant. Ultimately, it follows $\abs{\Delta \Theta} \sim (\sqrt{h}\alpha)^{-a}$ for \textit{ReLU} and \textit{Softplus} activation functions, with $a < 2$ and $a \to 2$ as depth grows. We provide \mm{scaling} arguments supporting these findings.
\end{abstract}

\section{Introduction and related works}\label{sec:intro}

Deep neural networks are successful at a variety of \mm{tasks}, yet understanding why they work remains a challenge. 
A surprising observation is that their performance on supervised tasks keeps increasing with their width $h$ in the over-parametrized  regime where they already fit all the training data \cite{neyshabur2017geometry,bansal2018minnorm,advani2017high,Spigler18}. This fact underlines the importance of describing deep learning in the limit $h\rightarrow \infty$.


The transmission of the signal through the network in the infinite-width limit is well understood at initialization. If the network weights  are initialized as i.i.d. random variables with zero mean and a fixed variance of order $h^{-\nicefrac12}$, the output function $f(x)$ is a Gaussian random processes with some covariance that can be computed~\cite{Neal1996,williams1997computing,Lee2017,matthews2018gaussian,novak2018bayesian,yang2019scaling}. 

Until recently much less was know about the \emph{evolution} of infinitely-wide networks. It turns out that two distinct limits emerge, depending on the initialization of the last layer of weights. 

\paragraph{Mean Field limit}
\mm{A limiting behavior of neural networks}, called ``mean field'' in the literature, has been studied in several works focusing mostly on one-hidden layer networks~\cite{mei2018mean,rotskoff2018neural,chizat2018,sirignano2018mean,mei2019mean,nguyen2019mean}. In this setting the output function of the network with $h$ hidden neurons corresponding to an input $x$ is
\begin{equation}
\label{111}
f(w,x) = \frac1h \sum_{i=1}^h c_i \sigma(a_i\cdot x + b_i),
\end{equation}
where $\sigma(\cdot)$ is the non-linear activation function and $w_i=(a_i, b_i, c_i)$ are the parameters associated with a hidden neuron. At initialization, the terms in the sum are independent random variables. We can invoke the law of large numbers: for large $h$ the average tends to the expectation value
\begin{equation}
f(w,x) \to \int\mathrm{d}a \mathrm{d}b \mathrm{d}c\, \rho(a, b, c)\; c \sigma(a \cdot x + b),
\end{equation}
and it has been shown in the literature that the training dynamics is controlled by a differential equation for the density of parameters $\rho$:
\begin{gather}
\partial_t \rho_t = 2 \nabla \cdot (\rho_t \nabla \Psi(a,b,c; \rho_t)),\\
\Psi(a,b,c; \rho) = V(a, b, c) + \int\mathrm{d}a^\prime \mathrm{d}b^\prime \mathrm{d}c^\prime\, \rho(a^\prime, b^\prime, c^\prime)\; U(a,b,c; a^\prime,b^\prime,c^\prime).
\end{gather}
Here $\nabla$ is the gradient with respect to $(a, b, c)$ and $U, V$ are potentials defined in~\cite{mei2018mean}.
This is equivalent to the hydrodynamic (continuous) description of interacting particles in some external potential. The performance of a network is then expected to plateau on approaching this limit, as $h\to\infty$.

\paragraph{NTK limit}
\mm{Another limit has been \mm{identified} when the weights in the last layer scale as $h^{-\nicefrac12}$, which differs from the $h^{-1}$ scaling used in the mean-field setting. This limit also applies to deep fully-connected networks and other architectures.}
The learning dynamics in this limit simplifies~\cite{jacot2018neural,Du2019,Allen-Zhu2018,lee2019wide,arora2019exact,park2019effect} and is entirely described by the \emph{neural tangent kernel} (NTK) defined as:
\begin{equation}
\Theta(w,x_1,x_2) = \nabla_w f(w,x_1) \cdot \nabla_w f(w,x_2),\label{eq:ntkformula}
\end{equation}
where $x_1,x_2$ are two inputs and  $\nabla_w$ is the gradient with respect to the parameters $w$. The kernel $\Theta$ is defined at any time and typically evolves during the dynamics. As $h\to\infty$, $\Theta(w,x_1,x_2)\to\Theta_\infty(x_1,x_2)$ does not vary at initialization (and thus it does not depend on the specific choice of $w$) and does not evolve in time: the kernel is frozen. The dynamics is guaranteed to converge on a time independent of $h$ to a global minimum of the loss, and as for usual kernel learning the function only evolves in the space spanned by the functions $\Theta_\infty(x_\mu,x)$, where $\{x_\mu, \mu=1...n\}$ is the training set.\vspace{1em}

The existence of two distinct limits raises both fundamental and practical questions. First, which limit best characterizes the neural networks that are used in practice, and which one leads to a better performance? These questions are still debated. In~\cite{chizat19}, based on teacher-student numerical experiments, it was argued that the NTK limit is unlikely to explain the success of neural networks. However, it was recently shown that the NTK limit is able to achieve good performance on real datasets~\cite{arora2019exact} (significantly better than alternative  kernel methods). Moreover, some predictions of the NTK limit agree with observations in networks of sizes that are used in practice~\cite{lee2019wide}. Second, it was argued that in the NTK limit the surprising improvement of performance with $h$ stems from the $h^{-\nicefrac12}$ fluctuations of the kernel at initialization, that ultimately leads to similar fluctuations in the learned function and degrades the performance of networks of small width~\cite{geiger2019scaling}. These fluctuations can be removed by ensemble-averaging output functions obtained with different initial conditions and trained independently, leading to an excellent performance already near the underparametrized to overparametrized (or \emph{jamming}) transition $h^*$ beyond which all the training data are correctly fitted~\cite{Geiger18}. Does this line of thought hold true in the mean-field limit?  

Finally, unlike in the NTK limit where preactivations vary very weakly during training, in the mean field limit feature training occurs. It is equivalent to saying that the tangent kernel (which can always be defined) evolves in time~\cite{rotskoff2018neural,mei2019mean,chizat19}. What are the characteristic time scales and magnitude of this evolution?

\subsection{Our contributions}
In this work we answer these questions empirically by learning a model of the form:
\begin{equation}
F(w,x) \equiv \alpha\left[f(w,x) - f(w_0,x)\right],
\label{eq:bachmodel}
\end{equation}
where $f(w,x)$ is a deep network \st{and $w_0$ is the network parameters at initialization}. This model is inspired by~\cite{chizat19}. For $\alpha=\mc O(1)$ it falls into the framework of the NTK literature, whereas for $\alpha=\mc O(h^{-\nicefrac12})$ it corresponds to the mean-field framework.
Our strategy is to consider a specific setup that is fast to train, simple and to some extent theoretically tractable. Thus we focus on 
fully-connected (FC) networks with \textit{Softplus} activation functions and gradient-flow dynamics. In the main body of the work we consider the Fashion-MNIST dataset, later in \sref{sec:other_setups} we extend our study to other datasets, architectures (including CNNs) and dynamics to verify the generality and robustness of our claims. For each setting we study systematically the role of both $\alpha$ (varied on 11 orders of magnitude) and the width $h$, learning ensembles of 10 to 20 networks so as to quantify precisely the magnitude of fluctuations induced by initialization and the benefit of ensemble averaging. 

Our main findings are as follows: (i) In the $(\alpha, h)$ plane two distinct regimes can be identified, separated by $\alpha^* = \mathcal{O}(h^{-\nicefrac12})$, in which performance and dynamics are qualitatively different. 
(ii) Which regime achieves a lower generalization error depends on the data structure and the network architecture. 
We find that fully-connected architectures generally perform better in the lazy-training regime, while convolutional networks trained on CIFAR10 with ADAM achieves a smaller error in the feature-training regime. 
(iii) The fluctuations of the network output $\delta F$ decay with the width as $\delta F = \mathcal{O}(h^{-\nicefrac12})$ in both regimes, leading to a performance that increases with the degree of overparametrization. Because of the nature of the fluctuations, a similar improvement can also be obtained at an intermediate width by ensemble-averaging several independently-trained networks. 
(iv) In the feature-training regime there exist a characteristic time scale $t_1 \sim \sqrt{h}\alpha$, such that for $t \ll t_1$ the dynamics is linear. 
At $t \sim t_1$, the output grows by a factor $\sqrt{h}$ and the tangent kernel $\abs{\Delta \Theta}$ varies significantly and can not be considered approximately constant any longer. Ultimately, it follows $\abs{\Delta \Theta} \sim (\sqrt{h}\alpha)^{-a}$ for \textit{ReLU} and \textit{Softplus} activation functions, with $a < 2$ and $a \to 2$ as depth grows. We provide scaling arguments supporting these findings.

The implications of our work are both practical (in terms of which parameters and architectures lead to improved performance) and conceptual (in quantifying the dynamics of feature training and providing informal explanations for these observations). More generally, it suggests that future empirical studies of deep learning would benefit from characterizing the regime in which they operate, since it will most likely impact their results.

The code used for this article is available online at \url{https://github.com/mariogeiger/feature_lazy/tree/article}.

\subsection{Related work}
Our work is most closely related to \cite{chizat19} and to \cite{chizat:hal-01945578} that appeared simultaneously to ours. 
The scale $\alpha \gg h^{-\nicefrac12}$ separating the lazy and feature learning regimes  was justified for a one-hidden layer,
in consistence with our findings that also apply to deep nets. The authors further suggest 
that the feature learning regime should outperform the lazy training one,  as they observe in two different setups: i) a fully-connected one-hidden-layer network trained on data from a teacher network and ii) a VGG-11 deep network  trained on CIFAR10 with a cross-entropy loss. Both setups achieve a smaller generalization error away from the lazy-training regime.
Our empirical analysis is much more extensive both in terms of dataset  and observables studied. It gives a different view,
since for real data we find that for fully-connected networks lazy training  tends to outperform feature learning.

\section{Notations and set-up} \label{sec:setup}

The following setup is used for all empirical results in the article, except for those presented in \sref{sec:other_setups}, where we explore different settings.

We consider deep networks with $L$ hidden layers performing a binary classification task. We denote by $w$ the set of parameters (or weights), and by $f(w,x)$ the output of a network parametrized by $w$ corresponding to an input pattern $x$. The set of training data is $\mc T = \{(x_\mu, y_\mu), \mu=1,\dots,n\}$, where $y_\mu=\pm1$ is the label associated with the pattern $x_\mu$ and $n$ is the number of training data. In what follows $\dot a$ is the notation we use for the time derivative of a variable $a$ during training.

\paragraph{Parameter $\alpha$:}
Instead of training a network $f(w,x)$, we train $F(w,x) = \alpha (f(w,x) - f(w_0,x))$,  {\it i.e} we use this quantity as our predictor and train the weights $w$ accordingly. Here $w_0$ is the network's parameters at initialization. In the over-parametrized regime, this functional form ensures that for $\alpha \to \infty$, we enter in what we call the lazy-training regime, where changes of weights are small. Indeed in order to obtain a zero loss $\alpha (f(w,x) - f(w_0,x))$ must be ${\cal O}(1)$, thus $1 \sim \alpha (f(w,x) - f(w_0,x)) \sim \alpha \nabla_w f(w_0, x) \cdot dw$. For large $\alpha$ then $|f(w,x) - f(w_0,x)|$ is small: the dynamics can hence be considered linear~\cite{chizat19}, with $f(w,x) - f(w_0,x) \approx \nabla_w f(w_0,x)\cdot(w-w_0)$. Since the gradient of $f(w_0,x)$ does not scale with $\alpha$, this implies that $\abs{dw} \equiv \abs{w - w_0} \sim \alpha^{-1}$.

Learning is achieved via the minimization of the loss function
\be
\mathcal{L}(w) = \frac{1}{\alpha^2 n} \sum_{(x,y) \in \mc T} \ell(\alpha (f(w, x) - f(w_0, x)), y),
\ee
where $\ell$ is the loss per pattern, \mm{$\mathcal{T}$ is the training set and $n = |\mathcal{T}|$}.
The prefactor $\alpha^{-2}$ ensures that the convergence time does not depend on $\alpha$ as $\alpha \rightarrow \infty$, since for that choice we have $\alpha \dot f(w_0) ={\cal O}(\alpha^0)$ \mm{(i.e. does not scale with $\alpha$ for large $\alpha$)}. 

Note that if we train directly $\alpha f(w,x)$ without removing its value at initialization, we expect no difference with the present setting for $\alpha$ of order one or much smaller. This statement is  confirmed numerically in \aref{sec:alphaf}. For  $\alpha \gg 1$ however, we find empirically that the learning dynamics does not converge.

\paragraph{Dynamics}
Since our goal is to build a connection between empirical and theoretical approaches, we focus on a discrete version
of continuous dynamics obeying simple differential equation. The simplest is the \emph{vanilla gradient descent} which reads $\dot w = - \nabla_w \mathcal L$ and does not depend on any hyper-parameters. 
This dynamics is run in a discretized form, with a time step that is adapted at each step to ensure that
\begin{gather}
\alpha \max_{x \in \mathcal{T}} |f(w_i, x) - f(w_{i+1}, x)| < 0.1, \\
\frac{\| \nabla_w \mathcal L_i - \nabla_w \mathcal L_{i+1} \|^2}{\|\nabla_w \mathcal L_i\| \; \|\nabla_w \mathcal L_{i+1}\|} < \epsilon_\nabla = 10^{-4}.
\label{eq:stepconstraint}
\end{gather}
In \aref{app:sanity_check} we checked that our results are independent of $\epsilon_\nabla$ for  $\epsilon_\nabla\leq 10^{-4}$.

\paragraph{Activation function}
The \textit{Softplus} activation function is defined as $\mathit{sp}_\beta(x) = \frac{1}{\beta} \ln(1 + e^{\beta x})$. The larger the parameter $\beta$, the sharper is the \textit{Softplus}, and as $\beta\to\infty$ it tends to a \emph{ReLU} activation function. In our experiments we take $\beta=5$.
In order to have preactivations of unit variance at initialization we also multiply the \textit{Softplus} by a prefactor $a \approx 1.404$. The activation function that we use at all hidden neurons is then
\begin{equation}
\sigma(x) = a \; \mathit{sp}_{\beta=5}(x)
\end{equation}

\begin{wrapfigure}[15]{r}{0.4\textwidth}
 \centering
 \vspace{-0.75cm}
 \hspace{-0.8cm}\scalebox{0.85}{\import{figures/}{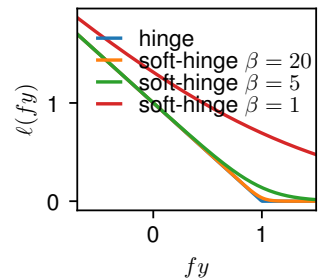}}\vspace{-0.6cm}
 \caption{Comparison between the hinge loss and the soft-hinge losses $\mathit{sp}_\beta(1 - fy)$ with $\beta=1, 5, 20$. As $\beta$ increases, the soft-hinge loss tends to the hinge loss.}
 \label{fig:loss}
\end{wrapfigure}

Non-smooth activation functions like ReLU can introduce additional phenomena; for instance, we quantify its impact on the evolution of the neural tangent kernel in \aref{app:ReLU}.

\paragraph{Loss function}
For the loss-per-pattern $\ell(f, y)$  we use the soft-hinge loss $\ell(f, y) = \mathit{sp}_\beta(1 - fy)$, where $\mathit{sp_\beta}$ is again a \textit{Softplus} function. This function is a smoothed version of the hinge loss, to which it tends as $\beta \to \infty$. We use the value $\beta=20$ as a compromise between smoothness and being similar to the hinge loss, see \fref{fig:loss}. The stopping criterion to end the learning dynamics is met when all patterns are classified within a sufficient margin, that is when $\alpha \left[f(w, x_\mu) - f(w_0, x_\mu)\right] y_\mu > 1$ for all $\mu=1,\dots, n$. Keep in mind that this criterion makes sense only in the overparametrized phase (as it will be the case in what follows), where networks manage to fit all the training set~\cite{Spigler18}.
The hinge loss results in essentially identical performance to the commonly used cross-entropy in state-of-the-art architectures, but leads to a dynamics that stops in a finite time in the over-parametrized regime considered here, removing the need to introduce an arbitrary temporal cut-off~\cite{Spigler18}.

\begin{figure}[ht]
    \centering
    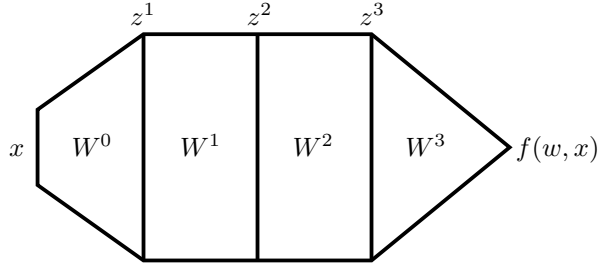
    \caption{Fully-connected architecture with $L=3$ hidden layers.
    \label{fig:network}}
\end{figure}

\paragraph{Architecture}
We use a constant-width fully-connected architecture based on~\cite{jacot2018neural}.
Given an input pattern $x \in \mathbb{R}^d$, we denote by $\tilde z^\ell$ the vector of preactivations at each hidden layer and by $z^\ell$ the corresponding activations. The flow of signals through the network can be written iteratively as
\begin{align}
&z^\ell = \sigma(\tilde z^\ell),\\
&\tilde z^{1} = d^{-\nicefrac12} W^0 x,\\
&\tilde z^{\ell+1} = h^{-\nicefrac12} W^\ell z^\ell,\\
&f(w, x) = h^{-\nicefrac12} W^L z^L.
\end{align}
The matrices $W^\ell$ contain the parameters of the $\ell$-th hidden layer, and the (vectorized) set of all these matrices has been previously denoted as $w$.
The width and depth of the network are $h$ and $L$; in our simulations we vary $h$ but we mostly keep the depth $L=3$ constant. The input patterns live a $d$-dimensional space. \fref{fig:network} illustrates the architecture and explains our notation. 
In order to have preactivations of order $\mathcal{O}(1)$ at initialization, all the weights are initialized as standard Gaussian random variables, $W^\ell_{ij} \sim \mathcal{N}(0,1)$.
Note that there is no bias, we discuss it in \aref{sec:bias}.

\paragraph{Dataset}
We train our network to classify $28\times28$ grayscale images of clothes from the Fashion-MNIST database \cite{xiao2017/online}.
For simplicity we split the original 10 classes in two sets and we perform binary classification.
For $(x, y) \in \mathcal{T}$ we have $x \in \mathbb R^{28\times28}$ ($d=784$) and $y = \pm 1$. The input is normalized on the sphere, $\sum_i x_i^2 = d$ such that each component $x_i$ has unit variance.
We took 1000 images of each class (10000 in total) to make our train set and 5000 of each class (50000 in total) to make our test set.
The train set is smaller than usual (10000 instead of 50000) in order to shorten the training time.

\section{Disentangling feature training and lazy training according to  performance}\label{sec:tworegimes}
\begin{figure}[th]
    \centering
    \scalebox{0.8}{\import{figures/}{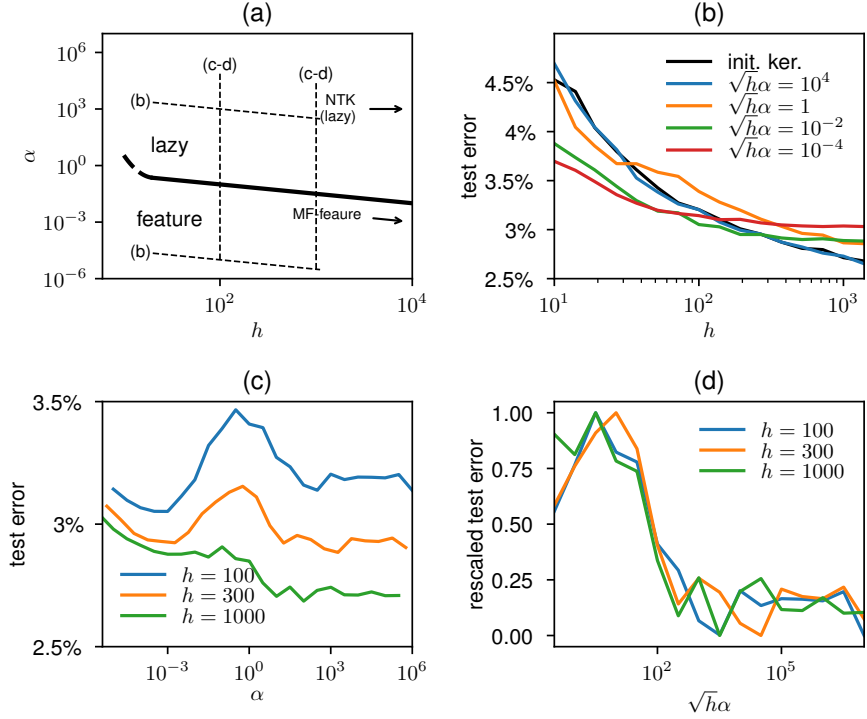}}
    \caption{\emph{(a)} Schematic representation of the parameters that we probe: either we fix $\alpha \sqrt{h}$ or we keep the width $h$ constant and we vary $\alpha$. The location of the cross-over between the lazy and feature-training regimes is also indicated. 
    \emph{(b)} Test error \textit{v.s.}  network's width $h$ for different values of $\sqrt{h} \alpha$ as indicated in legend. The black solid line is the test error of the frozen NTK at initialization, a limit that is recovered as $\alpha\sqrt{h}\to\infty$. \emph{(c)}
     Test error \textit{v.s.} $\alpha$, for different widths $h$. (averaged over 20 initializations) \emph{(d)} Same data as in \emph{(c)}: after an arbitrary affine transformation ($a x + b$) of the test error, the curves collapse when plotted against $\sqrt{h}\alpha$. 
     \label{fig:vary_alpha}}
\end{figure}

To argue the existence of two distinct regimes in deep neural networks, we evaluate their performance in the $(\alpha,h)$ plane. How we vary parameters to probe this plane  is represented in \fref{fig:vary_alpha} \emph{(a)}. We consider plots obtained at fixed $h$ and varying $\alpha$ (\fref{fig:vary_alpha} \emph{(c,d)}) as well as fixed $\sqrt{h}\alpha$ (\fref{fig:vary_alpha} \emph{(b)}).



\fref{fig:vary_alpha} \emph{(b)} shows the test error as a function of the width $h$, for different values of $\sqrt{h} \alpha$. For large $h$, we observe that as $\sqrt{h} \alpha$ is increased the performance also increases, up to a point where it converges to a limiting curve (for $\sqrt{h}\alpha\geq 10^3$ in the figures). This limiting curve coincides with the test error found if the NTK is frozen at initialization (see \aref{app:frozen_ntk} for a description of our dynamics in that case), represented by a black solid line. \mm{Following \cite{chizat19}}, we refer to this limiting behavior as lazy training. 

There is a value of $\alpha$ for which we leave the lazy-training regime and enter a nonlinear regime that we call feature training.\footnote{\mm{In our parlance, the NTK and mean-field limits correspond respectively to the lazy-training and feature-training regimes in the $h\to\infty$ limit.}}
To identify this regime, in \fref{fig:vary_alpha} \emph{(c)} we show the test error as a function of $\alpha$ for several widths $h$, and the rescaled test error (in such a way that it takes values between $0$ and $1$) as a function  of $\sqrt{h}\alpha$ in \fref{fig:vary_alpha} \emph{(d)}. The curves collapse, supporting that in the $(\alpha,h)$ plane the boundary between the two regimes lies at a scale $\alpha^* = \mc O(h^{-\nicefrac12})$. It is precisely the scaling used in Mean Field.

\section{Fluctuations of the output function and the effect of ensemble averaging}\label{sec:fluctuations}

\begin{figure}[th!]
    \centering
    \scalebox{0.8}{\import{figures/}{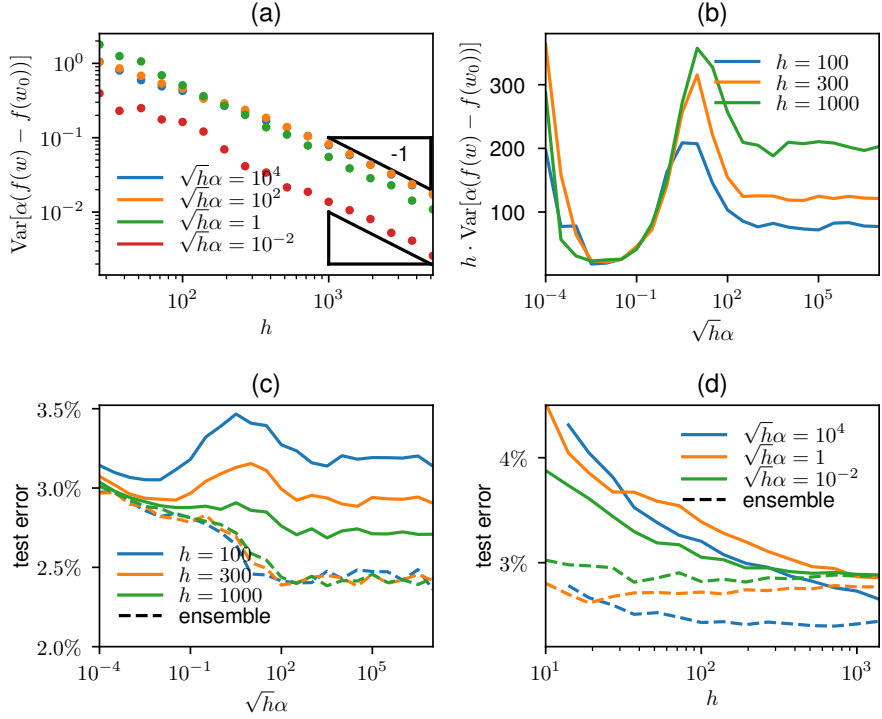}}
    \caption{\emph{(a)} Variance of the network's output \textit{v.s.} its width $h$, for different values of $\sqrt{h}\alpha$. In both regimes the variance scales as $\mr{Var}[ \alpha(f-f_0)] \sim h^{-1}$ (20 initializations per point). For this panel (a) alone, the size of the dataset was reduced to $10^3$ because $10^4$ requires higher $h$ to observe the asymptotic behavior. See \aref{app:var_p} for an analysis of the dependence in the size of the dataset. \emph{(b)} Variance of the network's output times its width \textit{v.s.} $\sqrt{h}\alpha$, for different network's width. In the lazy-training regime, it needs a larger $h$ (or a smaller $n$) to observe the overlap of the curves. \emph{(c)} Test error and ensemble-averaged test error \textit{v.s.} $\sqrt{h}\alpha$, for several widths $h$ (20 initializations per point). \emph{(d)} Test error and ensemble-averaged test error \textit{v.s.} $h$ at fixed values of $\sqrt{h}\alpha$ (20 initializations per point). Once ensemble averaged, lazy training performs better than feature training \label{fig:var_ens}}
\end{figure}

As shown in \fref{fig:vary_alpha}, performance increases with $h$: adding more trainable parameters leads to better predictability and deep networks do not overfit. This surprising behavior was related to the fluctuations of the output function  induced by initial conditions, observed to decrease with $h$ \cite{neal2018modern,geiger2019scaling}. To quantify the fluctuations in both regimes we train an ensemble of 20 identical functions $F(w_i,x) = \alpha \left[ f(w_i,x) - f(w_{i,0},x) \right]$ starting from different initial conditions, and then we measure the ensemble average $\bar F(x)$:
\begin{equation}
\bar F(x) \equiv \frac\alpha{20} \sum_{i=1}^{20} (f(w_i,x) - f(w_{i,0},x))
\end{equation}

A single trained realization of the network fluctuates around the ensemble average $\bar F(x)$, and therefore $\delta F(w,x) \equiv F(w,x) - \bar F(x)$ is a random function, whose fluctuations are quantified by the variance: 
\be
\mr{Var}\, F(w,x) = \left\langle \left[F(w_i,x_\mu) - \bar F(x_\mu)\right]^2 \right\rangle_{\substack{ \mu\in\mr{test} \\ i\in\mr{ensemble} }}\!\!\!,
\ee
where the average is both over the 20 output functions and over all points $x_\mu$ in a test set. Other norms can be used to quantify this variance, all yielding the same picture. In \fref{fig:var_ens} \emph{(a-b)} we compute these fluctuations either in the feature-training or in the lazy-training regimes. In both cases we find the same decay with the network width:
\be
\mr{Var} \; \alpha\left[ f(w,x) - f(w_0,x) \right] \sim h^{-1}. \label{eq:fluctuations}
\ee
In the feature-training regime, this observation is consistent with the predictions from~\cite{rotskoff2018neural} (obtained for a one-hidden layer performing regression).
Since $\delta F$ is a random function with zero expectation value and finite variance $\mr{Var}\, F \sim h^{-1}$ we will write that $\delta F \sim h^{-\nicefrac12}$.

In \fref{fig:var_ens} \emph{(b)}, in the lazy-training regime, the curves do not overlap. This is a preasymptotic effect. In \aref{app:var_p} we show that the asymptotic power law is reached for smaller dataset sizes.

It was argued in~\cite{geiger2019scaling} that in the lazy-training regime this scaling simply stems from the fluctuations of the NTK at initialization, that go as $\abs{\delta \Theta} \sim 1/\sqrt{h}$ and lead to similar fluctuations in $\delta F$. These fluctuations were argued to lead to an asymptotic decrease of test error as $1/h$, consistent with observations. Interestingly, the same scaling for the fluctuations holds in the feature-training regime, presumably reflecting the approximations expected from the Central Limit Theorem (CLT) when \eref{111} is replaced by an integral.

As a consequence of these fluctuations, ensemble averaging output functions leads to an enhanced performance in both regimes, as shown in \fref{fig:var_ens} \emph{(c-d)}. We remark that:

(i) In each regime, the test error of the ensemble average is essentially independent of $h$. It implies that the variation of performance with $h$ is only a matter of diminishing fluctuations in the over-parametrized case considered here (as shown below, we always fit all training data in these runs). It also supports that the plateau value of the ensemble-average  performance we observe corresponds to the performance of single network in the $h\rightarrow\infty$ limit. 

(ii) Interestingly, it is reported in \cite{geiger2019scaling} that for a fixed $\alpha$, the smallest test error of the ensemble average is obtained at some finite $h_{\min{}}$ beyond $h^*$ implying that past $h_{\min{}}$ performance is decreasing with growing $h$. It can now be simply explained: at fixed $\alpha$ one goes from feature to lazy training as $h$ increases, since $\sqrt{h}\alpha$ also increases and must eventually become much larger than one, leading to a change in performance.

(iii) For small values of $\sqrt{h} \alpha$ (smaller than $10^{-4}$) the variance blows up. We leave the study of this regime for future works. Since we observe that the test error also greatly increases for these values of $\sqrt{h} \alpha$, this regime is of less interest.

\section{Training dynamics differs in the two regimes}\label{sec:dynamics}

\begin{figure}[th!]
    \centering
    \scalebox{0.8}{\import{figures/}{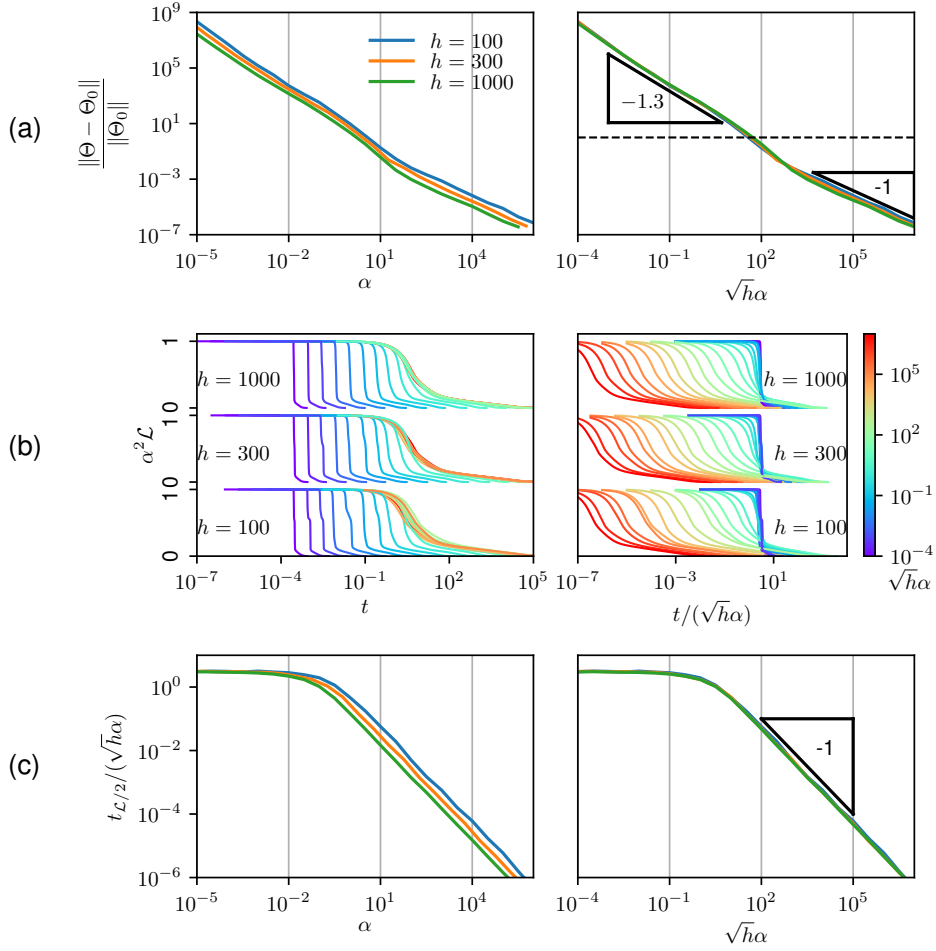}}
    \caption{\emph{(a)} Relative evolution of the kernel $\abs{\Theta(w) - \Theta(w_0)}/\abs{\Theta(w_0)}$ {\it v.s.} $\alpha$ (left) and $\sqrt{h} \alpha$ (right).
    \emph{(b)} Rescaled loss $\alpha^2{\cal L}$ {\it v.s.} $t$ (left) and $t/\sqrt{h}\alpha$ (right), for three different network's width and for different values of $\sqrt{h}\alpha$ indicated by the color bar.
    \emph{(c)} The time for which the loss is reduced by half \mm{(divided by $\sqrt{h}\alpha$)} {\it v.s.} $\alpha$ (left) and $\sqrt{h} \alpha$ (right). \mm{Notice that it reaches a plateau in the feature-training regime.}
    For \emph{(a)} and \emph{(c)} each point is averaged over 10 initializations.
    \label{fig:time_dynamics}}
\end{figure}

The network is able to learn features in the feature-training regime, while it cannot in the lazy-training regime because the NTK is frozen.
To quantify feature training we measure the total relative variation of the kernel at the end of training: $\abs{\Theta(w) - \Theta(w_0)} / \abs{\Theta(w_0)}$, where the norm of a kernel is defined as
$\abs{\Theta(w)}^2 = \sum_{\mu,\nu\in\mr{test\ set}} \Theta(w, x_\mu, x_\nu)^2$.
This quantity is plotted in \fref{fig:time_dynamics} \emph{(a, left)} versus $\alpha$ for several widths $h$, and versus $\sqrt{h}\alpha$ in \fref{fig:time_dynamics} \emph{(a, right)}. The fact that the curves collapse  indicates that $\alpha\sqrt{h}$ is the parameter that controls feature training. In particular, the crossover $\alpha^*\sim h^{-\nicefrac{1}{2}}$ precisely corresponds to the point where the change of the kernel is of the order of the norm of the kernel at initialization.
Moreover, in the feature-training regime we find:
\be
\label{ker}
\abs{\Theta(w) - \Theta(w_0)}/\abs{\Theta(w_0)} \sim (\sqrt{h}\alpha)^{-a},
\ee
with an exponent $a\approx 1.3$. By contrast in the lazy-training regime $\abs{\Theta(w) - \Theta(w_0)}/\abs{\Theta(w_0)} \sim 1/(\sqrt{h}\alpha)$, as expected for what concerns the dependency in $h$~\cite{jacot2019hessian,geiger2019scaling,lee2019wide}. 
In \aref{app:ReLU}, we show that for a non-smooth activation function like the ReLU there is a third intermediary regime where $\abs{\Theta(w) - \Theta(w_0)}/\abs{\Theta(w_0)} \sim \alpha^{-1/2}$.

In \aref{app:frozen_ntk}, we show in \fref{fig:kernel_dyn} an example of the evolution of the kernel (represented by its Gram matrix).
We also present an interesting discovery: once the network has been trained, performing kernel learning with the NTK obtained at the end essentially leads to the same generalization error.

We finally investigate the temporal evolution of learning, known to be characterized by several time scales \cite{Baity18}. The rescaled loss $\alpha^2{\cal L}$ is shown in \fref{fig:time_dynamics} \emph{(b)}. As expected \cite{jacot2018neural,chizat19}, in the lazy-training regime $\alpha^2 {\cal L}(t)$ depends neither on $\alpha$ nor $h$. This is not true in feature training however. \mm{We define $t_{\mathcal{L}/2}$ as the time for which the loss reduced by a half. Our key finding, visible in \fref{fig:time_dynamics} \emph{(b)}, is that the learning curves are very sharp in the feature-training regime and they overlap when we rescale the time axis properly. \fref{fig:time_dynamics} \emph{(b,c)} show that in the feature-training regime:}
\be
\label{time}
t_{\mathcal{L}/2}\sim \sqrt{h}\alpha.
\ee
\mm{In \sref{app:informalargument} we show that there is a time $t_1$ that marks the timescale below which the dynamics remains linear. In the feature-training regime we explain that $t_{\mathcal{L}/2}$ correspond to $t_1$ due to the nonlinearity that makes the dynamics evolve rapidly near $t_1$.}

\section{Arguments on kernel dynamics and regimes' boundary}\label{app:informalargument}

The arguments proposed in this section do not have the status of mathematical proofs. 
They provide heuristic explanations for the scaling  $t_1\sim \alpha \sqrt{h}$ and  $\alpha^*\sim 1/\sqrt{h}$,  and support that the output function grows by a factor $\sqrt{h}$ before the dynamics become highly non-linear. A simple assumption on the nature of the ensuing non-linear dynamics leads to the prediction \eref{ker} with $a=2/(1+1/L)$, which we view as a good approximation (not necessarily exact) of our observations. 

Our starting point is that for the NTK initialization, we have at $t=0$ that $\tilde z_\alpha={\cal O}(1)$ and $\partial f/\partial \tilde z_\alpha= {\cal O}(1/\sqrt{h})$,
where $\tilde z_\alpha$ is the preactivation of a hidden neuron.  The second point can be derived iteratively starting from the last hidden layer. 
Denote by $W^0$ the weight matrix connected to the input, and $W^\ell$ the weight matrix connecting two hidden neurons in the $\ell-1$ and $\ell$ hidden layers respectively, and $W^L$ the weight vector connected to the output. It is then straightforward to show using the chain rule and $\sigma'(\tilde z)={\cal O}(1)$ that (see also \cite{arora2019exact}):
\be
\label{aa1}
\frac{\partial f}{\partial W^0}={\mathcal{O}}\left(\frac{1}{\sqrt{h}}\right);\quad
\frac{\partial f}{\partial W^\ell}={\mathcal{O}}\left(\frac{1}{h}\right);\quad
\frac{\partial f}{\partial W^L}={\mathcal{O}}\left(\frac{1}{\sqrt{h}}\right).
\ee

From which we deduce that:
\be
\label{aa2}
\dot W^0={\mathcal{O}}\left(\frac{1}{\sqrt{h} \alpha}\right);\quad
\dot W^\ell={\mathcal{O}}\left(\frac{1}{h \alpha}\right);\quad
\dot W^L={\mathcal{O}}\left(\frac{1}{\sqrt{h}\alpha}\right)
\ee
\mm{by using the gradient-descent formula (i.e. $\dot w = -\nabla \mathcal{L}$).} 

Next, we consider how the neurons' preactivations evolve in time. From the composition of derivatives, one obtains $\dot{ \tilde z}^{\ell+1} = h^{-\nicefrac12} (\dot W^\ell z^\ell + W^\ell \dot z^\ell)$.
It is clear from \eref{aa2} that $h^{-\nicefrac12} \dot W^\ell z^\ell={\mathcal{O}}\left(\frac{1}{\sqrt{h}\alpha}\right)$.  Concerning the product $W^\ell \dot z^\ell$, in the large-width limit it can be proven to be correctly estimated by considering that $W^\ell$ and $\dot z^\ell$ are independent \cite{Dyer19}. (This result simply stems from the fact that the time-derivative of the preactivation of one neuron depends on all its $h$ outgoing  weights, and is therefore weakly correlated to any of them). From the central limit theorem, the vector $W^\ell \dot z^\ell$ is thus of order $\sqrt{h}  \dot z^\ell$. Proceeding recursively from the input to the output we obtain:
\be \label{aa21}
\dot {\tilde z}^\ell={\mathcal{O}}\left(\frac{1}{\sqrt{h} \alpha}\right).
\ee
We checked \eref{aa21} numerically in \aref{app:dot_z}.

From \eref{aa2} and \eref{aa21} we expect that:
\be
\label{aa3}
\forall t \ll t_1 \equiv \alpha \sqrt{h}, \ \ W^L(t)-W^L(0)=o(1); \quad  \tilde{z}^\ell(t)-\tilde{z}^\ell(0)=o(1)
\ee
Thus for $t \ll t_1$, we are in the lazy-training regime where preactivations and weights did not have time to evolve, and we expect the kernel variations to be small
(see \cite{mei2019mean} for a related discussion). Since the lazy-training regime finds a zero loss solution and stops in a time ${\cal O}(1)$ (see \sref{sec:setup}),  if $t_1 = \alpha \sqrt{h} \gg 1$ the network remains in it throughout learning. 
Thus $\alpha^*\sim 1/\sqrt{h}$, as proposed for a single layer in \cite{chizat19}.

By contrast, if $\alpha \sqrt{h} \ll 1$ the dynamics has not stopped at  times $t \sim t_1$, for which  we have  $W^L(t_1)-W^L(0)={\cal O}(1)$ and $\tilde z(t_1)-\tilde z(0)={\cal O}(1)$: both the preactivations and the weights of the last layer  have changed significantly, leading to significant changes of $\nabla_w f$ and  $\Theta$. It is important to note that at $t\sim t_1$, the scale of the output function is expected to change.
Indeed at initialization the output  function, which is a sum made on the last layer of hidden neurons $f(w,x)= \frac{1}{\sqrt{h}} \sum_{i=1}^h W^L_i \sigma(\tilde z^L_i)$,  is ${\cal O}(1)$
as expected from the CLT applied to $h$ uncorrelated terms \cite{Neal1996}. However, for $t\sim t_1$ this independence does not hold anymore, since  the terms  $W^L_i \sigma(\tilde z^L_i)$ have evolved by ${\cal O}(1)$ to change the function $f(w,x)$ in a specific direction. We thus expect these correlations to build up linearly in time for $t\in[0,t_1]$ and to ultimately increase the output by a factor $\sqrt{h}$ at $t\sim t_1$, as confirmed in \aref{app:out}. Note that this effect does not appear at intermediate layers in the network, because the weights evolve much more slowly there as follows from \eref{aa2}.

Still, such an increase in the output is insufficient to find solutions deep in the feature-training regime, since $\alpha [f(w(t_1))-f(w(0))]\sim \alpha \sqrt{h} \ll 1$. We propose that for activation functions that increase linearly at large arguments as those we use, the dynamics for $t \approx t_1$ approximately corresponds to an inflation of the weights  along the direction $\dot w(t_1)$. Specifically, we define an amplification factor $\lambda(t) = \abs{w(t)-w(0)}/\abs{ w(t_1)-w(0)}$, and assume for simplicity that this amplification is identical in each of the $L+1$ layers of weights (we disregard in particular the fact that the last and first layer may behave differently, as discussed in \cite{arora2019exact}).  By definition, $\lambda(t_1)=1$. At the end $t_2$ of training,  for activation functions that increase linearly at large arguments, we expect to have  $\lambda(t_2)\sim [\alpha \sqrt{h}]^{-1/(L+1)}$  to ensure that $\alpha [f(w(t_2))-f(w(0))]={\cal O}(1)$. Gradient with respect to weights are increased by $\lambda^L$, leading to an overall inflation of the kernel:
\be
\label{aa4}
\Theta(t_2)-\Theta(0)\sim \Theta(t_2)\sim \lambda^{2L}\sim [\alpha \sqrt{h}]^{-\frac{2}{1+1/L}}
\ee
leading to $a=1.66$ consistent with \eref{ker}. We have checked  this prediction with success for shallow networks with $L=2$, as shown in \aref{app:shallow}.

\begin{table}[b]
\centering
\caption{\label{tab:perf}Performance for different setups.}
\begin{tabular}{|l|l|l|l|} 
\hline
Architecture & Dataset (binary, 10k) & Algorithm & Regime performing better \\
\hline
CNN (4 hidden layers) & CIFAR10 & ADAM (batch size 32) & feature training \\
CNN (4 hidden layers) & Fashion-MNIST & ADAM (batch size 32) & Not clear (\fref{fig:other}(b)) \\
FC (3 hidden layers) & CIFAR10 & ADAM (batch size 32) & lazy training \\
FC (3,9 hidden layers) & Fashion-MNIST & Gradient flow & lazy training \\
FC (3 hidden layers) & MNIST & Gradient flow & lazy training \\
FC (3 hidden layers) & EMNIST letters & Gradient flow & lazy training \\
FC (3 hidden layers) & CIFAR10 & Gradient flow & lazy training \\
FC (5 hidden layers) & MNIST 10 PCA & Gradient flow & feature training \\
\hline
\end{tabular}
\end{table}

\section{Other experiments}
\label{sec:other_setups}

We now check that our conclusions extend to other data sets, architectures and learning dynamics. In particular we consider under which circumstances feature training outperforms lazy training.  Table~\ref{tab:perf} summarizes our results. The key observation is that which regime works best depends on the architecture and on the data. In particular, for the training set size $10^4$ we focus on in this study (as it allows to study gradient descent in a reasonable time), we generally find that FC performs better under the lazy training regime. For the CNN architectures \mm{that we study, feature} training tends to perform better.   

\begin{figure}[th!]
    \centering
    \scalebox{0.8}{\import{figures/}{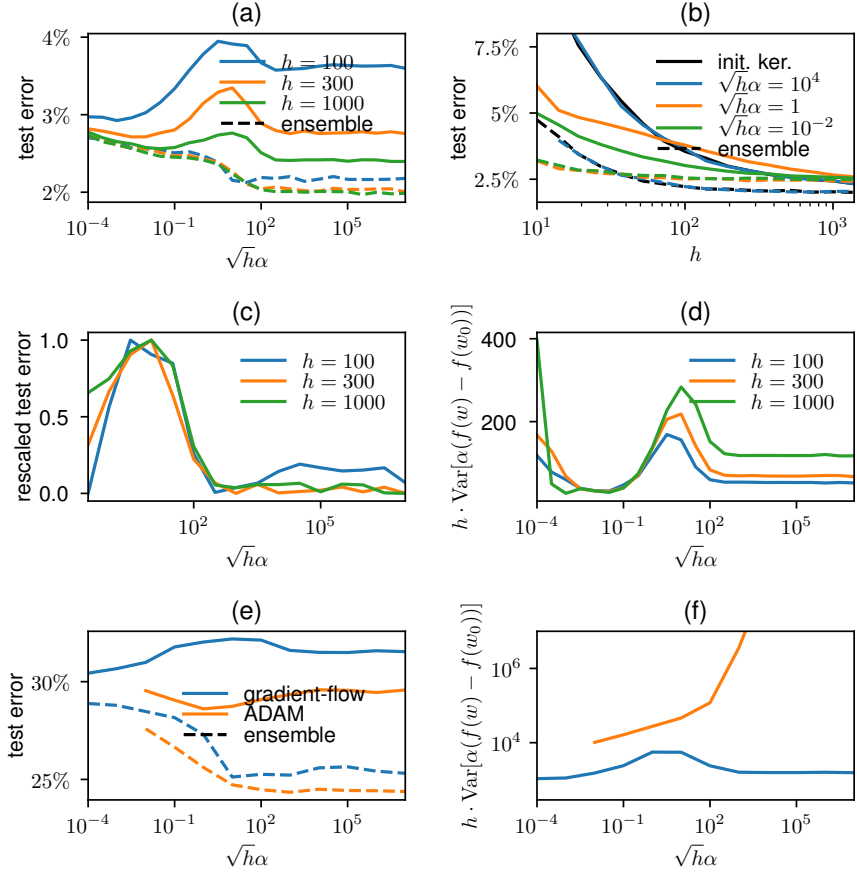}}\\
    \caption{\label{fig:mnistcifar}Binary classification on MNIST (\emph{a-d}) and CIFAR10 (\emph{e-f}) with the setup presented in \sref{sec:setup}. \emph{(a)} MNIST test error (single shot and ensemble average over 20 instances) \textit{v.s.} $\sqrt{h} \alpha$ for different widths $h$. \emph{(b)} MNIST test error \textit{v.s.} the width $h$ for different values of $\sqrt{h} \alpha$. The black lines is the test error of the frozen NTK at initialization, limit that is recovered as $\alpha \to \infty$. Ensemble averages are computed over 10 instances. \emph{(c)} Same data as in \emph{(a)}: after rescaling the test error to be in $(0,1)$ the curves collapse when plotted against $\sqrt{h}\alpha$. \emph{(d)} The network's width times the variance of the output \textit{v.s.} $\sqrt{h} \alpha$ for different widths $h$. This plot is computed for the MNIST dataset and is averaged over 20 initializations. \emph{(e)} CIFAR10 test error (single shot and ensemble average over 20 instances) \textit{v.s.} $\sqrt{h} \alpha$ for a network of width $h=100$. \emph{(f)} The network's width times the variance of the output \textit{v.s.} $\sqrt{h} \alpha$. This plot is computed for the CIFAR10 dataset and is averaged over 20 initializations.}
\end{figure}

\paragraph{MNIST and CIFAR10:} We train the FC network defined in \sref{sec:setup} on the MNIST and CIFAR10 datasets. CIFAR10 dataset contains $50000+10000$ $32\times32$ images in the trainset and testset, which are split into 10 classes. We reduced the trainset to 10000 images split into 2 classes (5 classes merges into 1) and the images are flatten into a vector of size 1024. In \fref{fig:mnistcifar} we show the results. The picture is qualitatively similar to what we see for Fashion-MNIST.

\begin{figure}[th!]
    \centering
    \scalebox{0.8}{\import{figures/}{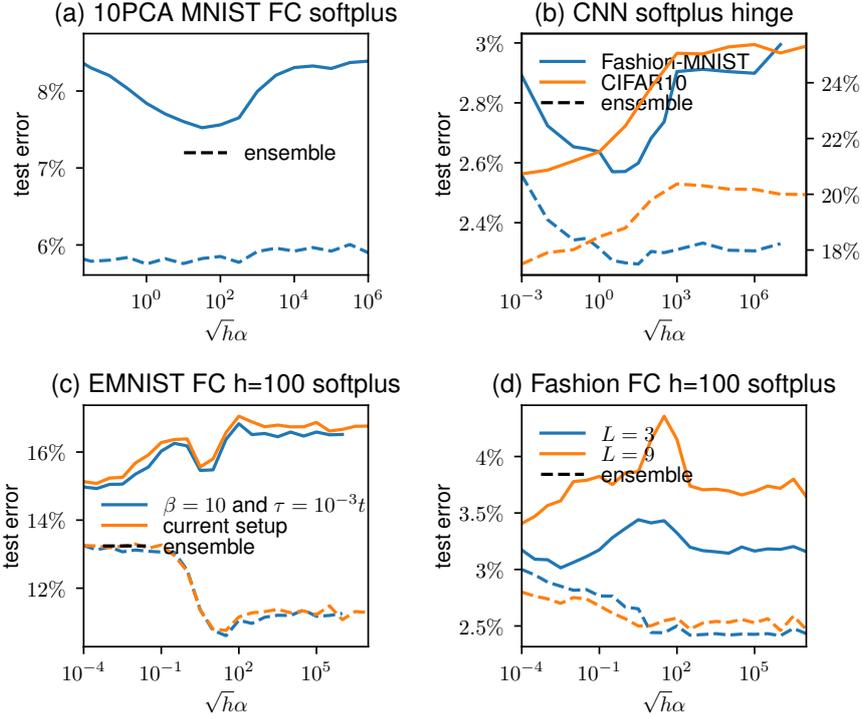}}
    \caption{Test error and ensemble average test error \textit{v.s.} $\sqrt{h} \alpha$ for different setup. \emph{(a)} MNIST reduced to its first 10 PCA component trained on FC of width 100. \emph{(b)} CIFAR10 trained on a CNN with the hinge loss and ADAM. \emph{(c)} The letters from EMNIST. \emph{(d)} Two different depth on Fashion-MNIST.\label{fig:other}}
\end{figure}

\paragraph{First ten principal components of MNIST PCA:}
As dataset we consider the projection of the handwritten digits in the MNIST dataset onto their first 10 principal components, obtained via principal-component analysis (PCA).
This dataset is harder to fit than the original MNIST. 
Using the FC network defined in \sref{sec:setup}, we observe (see \fref{fig:other} \emph{(a)}) that the test error for a single network  as well as the ensemble-averaged test error are nearly identical, even somewhat smaller in the feature-training regime than in the lazy-training regime, contrarily to what was observed for the full MNIST dataset.

\paragraph{EMNIST:}
Using again the FC network defined in \sref{sec:setup}, we consider the pictures of handwritten letters in the EMNIST dataset.
\fref{fig:other} \emph{(c)} shows that the results are similar to what observed for Fashion-MNIST and MNIST.
Here two setups are compared, the current setup (described in \sref{sec:setup}) and an alternative setup with $\beta=10$ for the loss function and a value of $\tau$ proportional to the time.

\paragraph{CNN for CIFAR 10 and Fashion-MNIST:}
We train a convolutional neural network (CNN) with 4 hidden layers (with stride and padding). It has \textit{Softplus} activation function and no biases. We initialize the parameters of the convolutional networks as standard Gaussians in such a way that the preactivations are of order unity.
Our architecture has 4 hidden layers, and we use a stride of \texttt{/2} before the first layer and in the middle of the network. After the convolutions we average the spatial dimensions, and the last layer is a simple perceptron. The code is available on the repository. It is trained to classify images of the CIFAR10 dataset ($32\times32$ images). Our trainset contains 10000 images split into 2 classes. Learning is achieved with the ADAM dynamics~\cite{Kingma14}.
We observe, see \fref{fig:other} \emph{(b)}, that individual and ensemble-averaged performance is better in the feature-training regime. For Fashion-MNIST, the optimal performance  occurs in a narrow range of intermediate $\alpha$.

\paragraph{Effect of Depth:}
In \fref{fig:other} \emph{(d)} we compare networks with different depths, $L=3$ with $L=9$. The setup is defined in \sref{sec:setup}.
We find that increasing depth does not change qualitatively the dependence of the test error with $\sqrt{h}\alpha$. Quantitatively, depth has a very limited effect on the ensemble average performance, but does decrease the performance of individual networks --- presumably indicating that depth increases fluctuations of the output function induced by random initialization.

\section{Conclusion}

We have shown that as the width $h$ and the output scale $\alpha$ at initialization are varied, two regimes appear depending on the value of $\alpha\sqrt{h}$. In the feature-training regime features are learned (in the sense that the tangent kernel evolves during training), whereas in the lazy-training regime the dynamics is controlled by a frozen kernel. Our key findings are that: (i) In both regimes, fluctuations induced by initialization decrease with $h$, explaining why performance increases with the width. (ii) In feature training, the learning dynamics is linear for $t \ll t_1 \sim \sqrt{h}\alpha$, time at which the output of the model becomes of order $\sqrt{h}\alpha$. If $\sqrt{h}\alpha \ll 1$, in order to fit the data the dynamics enters a non-linear regime for $t \sim t_1$ that affects the magnitude of the kernel.

In treating these data sets, we have separated classes randomly into two groups. In the future, it would be interesting to test performance with a binary classification where the categories are intuitively meaningful.
We have focused our comparison between these two learning regimes on a fixed data set size $p$, comparable in order of magnitude but smaller than the full data set. 
In a recent work\cite{paccolat2020geometric} we performed that comparison as $p$ varies for a CNN trained on MNIST. In that case, we found that for all $p$ considered, feature learning outperforms lazy training in terms of generalization error $\epsilon$. For the training curve $\epsilon \sim p^{-\beta}$  with $\beta_\text{lazy}=1/3$ and  $\beta_\text{feature} = 1/2$, indicting that the feature learning regime was improving faster with growing $p$ than lazy training in relative terms.

On the empirical side, our work supports that studies of deep learning (e.g. on the role of regularization) should specify in which regime their networks operate, since it is very likely that it affects their results. On the theoretical side, there is little quantitative understanding on how much the performance should differ in two regimes. The results that we presented in \sref{sec:other_setups} show that it depends on both the structure of the data and on the architecture of the network. Answering this point appears necessary to ultimately understand why deep learning works.

\subsection*{Acknowledgements}
We thank Levent Sagun, Cl\'ement Hongler, Franck Gabriel, Giulio Biroli, St\'ephane d'Ascoli for helpful discussions.
We thank Riccardo Ravasio and Jonas Paccolat for proofreading.
This work was partially supported by the grant from the Simons Foundation (\#454953 Matthieu Wyart). 
M.W. thanks the Swiss National Science Foundation for support under Grant No. 200021-165509.

\bibliographystyle{plainnat}
\bibliography{main}

\begin{thebibliography}{34}
\providecommand{\natexlab}[1]{#1}
\providecommand{\url}[1]{\texttt{#1}}
\expandafter\ifx\csname urlstyle\endcsname\relax
  \providecommand{\doi}[1]{doi: #1}\else
  \providecommand{\doi}{doi: \begingroup \urlstyle{rm}\Url}\fi

\bibitem[Advani and Saxe(2017)]{advani2017high}
Madhu~S Advani and Andrew~M Saxe.
\newblock High-dimensional dynamics of generalization error in neural networks.
\newblock \emph{arXiv preprint arXiv:1710.03667}, 2017.

\bibitem[Allen-Zhu et~al.(2018)Allen-Zhu, Li, and Song]{Allen-Zhu2018}
Zeyuan Allen-Zhu, Yuanzhi Li, and Zhao Song.
\newblock A convergence theory for deep learning via over-parameterization.
\newblock \emph{arXiv preprint arXiv:1811.03962}, 2018.

\bibitem[Arora et~al.(2019)Arora, Du, Hu, Li, Salakhutdinov, and
  Wang]{arora2019exact}
Sanjeev Arora, Simon~S Du, Wei Hu, Zhiyuan Li, Ruslan Salakhutdinov, and
  Ruosong Wang.
\newblock On exact computation with an infinitely wide neural net.
\newblock \emph{arXiv preprint arXiv:1904.11955}, 2019.

\bibitem[Baity-Jesi et~al.(2018)Baity-Jesi, Sagun, Geiger, Spigler, Arous,
  Cammarota, LeCun, Wyart, and Biroli]{Baity18}
Marco Baity-Jesi, Levent Sagun, Mario Geiger, Stefano Spigler, Gerard~Ben
  Arous, Chiara Cammarota, Yann LeCun, Matthieu Wyart, and Giulio Biroli.
\newblock Comparing dynamics: Deep neural networks versus glassy systems.
\newblock In Jennifer Dy and Andreas Krause, editors, \emph{Proceedings of the
  35th International Conference on Machine Learning}, volume~80 of
  \emph{Proceedings of Machine Learning Research}, pages 314--323,
  Stockholmsmässan, Stockholm Sweden, 10--15 Jul 2018. PMLR.
\newblock URL \url{http://proceedings.mlr.press/v80/baity-jesi18a.html}.

\bibitem[Bansal et~al.(2018)Bansal, Advani, Cox, and Saxe]{bansal2018minnorm}
Yamini Bansal, Madhu Advani, David~D Cox, and Andrew~M Saxe.
\newblock Minnorm training: an algorithm for training overcomplete deep neural
  networks.
\newblock \emph{arXiv preprint arXiv:1806.00730}, 2018.

\bibitem[Chizat and Bach(2018)]{chizat2018}
L\'{e}na\"{\i}c Chizat and Francis Bach.
\newblock {On the Global Convergence of Gradient Descent for Over-parameterized
  Models using Optimal Transport}.
\newblock In \emph{Advances in Neural Information Processing Systems 31}, pages
  3040--3050. Curran Associates, Inc., 2018.

\bibitem[Chizat and Bach(2019)]{chizat19}
Lenaic Chizat and Francis Bach.
\newblock {A Note on Lazy Training in Supervised Differentiable Programming}.
\newblock working paper or preprint, February 2019.
\newblock URL \url{https://hal.inria.fr/hal-01945578}.

\bibitem[Chizat et~al.(2019)Chizat, Oyallon, and Bach]{chizat:hal-01945578}
Lenaic Chizat, Edouard Oyallon, and Francis Bach.
\newblock {On Lazy Training in Differentiable Programming}.
\newblock In \emph{{NeurIPS 2019 - 33rd Conference on Neural Information
  Processing Systems}}, Vancouver, Canada, December 2019.
\newblock URL \url{https://hal.inria.fr/hal-01945578}.

\bibitem[de~G.~Matthews et~al.(2018)de~G.~Matthews, Hron, Rowland, Turner, and
  Ghahramani]{matthews2018gaussian}
Alexander~G. de~G.~Matthews, Jiri Hron, Mark Rowland, Richard~E. Turner, and
  Zoubin Ghahramani.
\newblock Gaussian process behaviour in wide deep neural networks.
\newblock In \emph{International Conference on Learning Representations}, 2018.
\newblock URL \url{https://openreview.net/forum?id=H1-nGgWC-}.

\bibitem[Du et~al.(2019)Du, Zhai, Poczos, and Singh]{Du2019}
Simon~S. Du, Xiyu Zhai, Barnabas Poczos, and Aarti Singh.
\newblock Gradient descent provably optimizes over-parameterized neural
  networks.
\newblock In \emph{International Conference on Learning Representations}, 2019.
\newblock URL \url{https://openreview.net/forum?id=S1eK3i09YQ}.

\bibitem[Dyer and Gur-Ari(2019)]{Dyer19}
Ethan Dyer and Guy Gur-Ari.
\newblock Asymptotics of wide networks from feynman diagrams.
\newblock \emph{arXiv preprint arXiv:1909.11304}, 2019.

\bibitem[Geiger et~al.(2018)Geiger, Spigler, d'Ascoli, Sagun, Baity-Jesi,
  Biroli, and Wyart]{Geiger18}
Mario Geiger, Stefano Spigler, {St{\'e}phane} d'Ascoli, Levent Sagun, Marco
  Baity-Jesi, Giulio Biroli, and Matthieu Wyart.
\newblock The jamming transition as a paradigm to understand the loss landscape
  of deep neural networks.
\newblock \emph{arXiv preprint arXiv:1809.09349}, 2018.

\bibitem[Geiger et~al.(2019)Geiger, Jacot, Spigler, Gabriel, Sagun, d'Ascoli,
  Biroli, Hongler, and Wyart]{geiger2019scaling}
Mario Geiger, Arthur Jacot, Stefano Spigler, Franck Gabriel, Levent Sagun,
  St{\'e}phane d'Ascoli, Giulio Biroli, Cl{\'e}ment Hongler, and Matthieu
  Wyart.
\newblock Scaling description of generalization with number of parameters in
  deep learning.
\newblock \emph{arXiv preprint arXiv:1901.01608}, 2019.

\bibitem[Jacot et~al.(2018)Jacot, Gabriel, and Hongler]{jacot2018neural}
Arthur Jacot, Franck Gabriel, and Cl{\'e}ment Hongler.
\newblock Neural tangent kernel: Convergence and generalization in neural
  networks.
\newblock In \emph{Proceedings of the 32Nd International Conference on Neural
  Information Processing Systems}, NIPS'18, pages 8580--8589, USA, 2018. Curran
  Associates Inc.
\newblock URL \url{http://dl.acm.org/citation.cfm?id=3327757.3327948}.

\bibitem[Jacot et~al.(2019)Jacot, Gabriel, and Hongler]{jacot2019hessian}
Arthur Jacot, Franck Gabriel, and Cl{\'e}ment Hongler.
\newblock The asymptotic spectrum of the hessian of dnn throughout training.
\newblock \emph{arXiv preprint arXiv:1910.02875}, 2019.

\bibitem[Kingma and Ba(2015)]{Kingma14}
Diederik~P Kingma and Jimmy Ba.
\newblock Adam: A method for stochastic optimization.
\newblock \emph{International Conference on Learning Representations}, 2015.

\bibitem[Lee et~al.(2018)Lee, Bahri, Novak, Schoenholz, Pennington, and
  Sohl-Dickstein]{Lee2017}
Jae~Hoon Lee, Yasaman Bahri, Roman Novak, Samuel~S. Schoenholz, Jeffrey
  Pennington, and Jascha Sohl-Dickstein.
\newblock Deep neural networks as gaussian processes.
\newblock \emph{ICLR}, 2018.

\bibitem[Lee et~al.(2019)Lee, Xiao, Schoenholz, Bahri, Sohl-Dickstein, and
  Pennington]{lee2019wide}
Jaehoon Lee, Lechao Xiao, Samuel~S Schoenholz, Yasaman Bahri, Jascha
  Sohl-Dickstein, and Jeffrey Pennington.
\newblock Wide neural networks of any depth evolve as linear models under
  gradient descent.
\newblock \emph{arXiv preprint arXiv:1902.06720}, 2019.

\bibitem[Mei et~al.(2018)Mei, Montanari, and Nguyen]{mei2018mean}
Song Mei, Andrea Montanari, and Phan-Minh Nguyen.
\newblock A mean field view of the landscape of two-layers neural networks.
\newblock \emph{arXiv preprint arXiv:1804.06561}, 2018.

\bibitem[Mei et~al.(2019)Mei, Misiakiewicz, and Montanari]{mei2019mean}
Song Mei, Theodor Misiakiewicz, and Andrea Montanari.
\newblock Mean-field theory of two-layers neural networks: dimension-free
  bounds and kernel limit.
\newblock \emph{arXiv preprint arXiv:1902.06015}, 2019.

\bibitem[Neal et~al.(2019)Neal, Mittal, Baratin, Tantia, Scicluna,
  Lacoste-Julien, and Mitliagkas]{neal2018modern}
Brady Neal, Sarthak Mittal, Aristide Baratin, Vinayak Tantia, Matthew Scicluna,
  Simon Lacoste-Julien, and Ioannis Mitliagkas.
\newblock A modern take on the bias-variance tradeoff in neural networks.
\newblock 2019.
\newblock URL \url{https://openreview.net/forum?id=HkgmzhC5F7}.

\bibitem[Neal(1996)]{Neal1996}
Radford~M. Neal.
\newblock \emph{Bayesian Learning for Neural Networks}.
\newblock Springer-Verlag New York, Inc., Secaucus, NJ, USA, 1996.
\newblock ISBN 0387947248.

\bibitem[Neyshabur et~al.(2017)Neyshabur, Tomioka, Salakhutdinov, and
  Srebro]{neyshabur2017geometry}
Behnam Neyshabur, Ryota Tomioka, Ruslan Salakhutdinov, and Nathan Srebro.
\newblock Geometry of optimization and implicit regularization in deep
  learning.
\newblock \emph{arXiv preprint arXiv:1705.03071}, 2017.

\bibitem[Nguyen(2019)]{nguyen2019mean}
Phan-Minh Nguyen.
\newblock Mean field limit of the learning dynamics of multilayer neural
  networks.
\newblock \emph{arXiv preprint arXiv:1902.02880}, 2019.

\bibitem[Novak et~al.(2019)Novak, Xiao, Bahri, Lee, Yang, Abolafia, Pennington,
  and Sohl-dickstein]{novak2018bayesian}
Roman Novak, Lechao Xiao, Yasaman Bahri, Jaehoon Lee, Greg Yang, Daniel~A.
  Abolafia, Jeffrey Pennington, and Jascha Sohl-dickstein.
\newblock Bayesian deep convolutional networks with many channels are gaussian
  processes.
\newblock In \emph{International Conference on Learning Representations}, 2019.
\newblock URL \url{https://openreview.net/forum?id=B1g30j0qF7}.

\bibitem[Paccolat et~al.(2020)Paccolat, Petrini, Geiger, Tyloo, and
  Wyart]{paccolat2020geometric}
Jonas Paccolat, Leonardo Petrini, Mario Geiger, Kevin Tyloo, and Matthieu
  Wyart.
\newblock Geometric compression of invariant manifolds in neural nets, 2020.

\bibitem[Park et~al.(2019)Park, Sohl-Dickstein, Le, and Smith]{park2019effect}
Daniel~S Park, Jascha Sohl-Dickstein, Quoc~V Le, and Samuel~L Smith.
\newblock The effect of network width on stochastic gradient descent and
  generalization: an empirical study.
\newblock \emph{arXiv preprint arXiv:1905.03776}, 2019.

\bibitem[Rotskoff and Vanden-Eijnden(2018)]{rotskoff2018neural}
Grant~M Rotskoff and Eric Vanden-Eijnden.
\newblock Neural networks as interacting particle systems: Asymptotic convexity
  of the loss landscape and universal scaling of the approximation error.
\newblock \emph{arXiv preprint arXiv:1805.00915}, 2018.

\bibitem[Sirignano and Spiliopoulos(2018)]{sirignano2018mean}
Justin Sirignano and Konstantinos Spiliopoulos.
\newblock Mean field analysis of neural networks.
\newblock \emph{arXiv preprint arXiv:1805.01053}, 2018.

\bibitem[Spigler et~al.(2018)Spigler, Geiger, d'Ascoli, Sagun, Biroli, and
  Wyart]{Spigler18}
Stefano Spigler, Mario Geiger, St{\'e}phane d'Ascoli, Levent Sagun, Giulio
  Biroli, and Matthieu Wyart.
\newblock A jamming transition from under-to over-parametrization affects loss
  landscape and generalization.
\newblock \emph{arXiv preprint arXiv:1810.09665}, 2018.

\bibitem[Williams(1997)]{williams1997computing}
Christopher~KI Williams.
\newblock Computing with infinite networks.
\newblock In \emph{Advances in neural information processing systems}, pages
  295--301, 1997.

\bibitem[Xiao et~al.(2017)Xiao, Rasul, and Vollgraf]{xiao2017/online}
Han Xiao, Kashif Rasul, and Roland Vollgraf.
\newblock Fashion-mnist: a novel image dataset for benchmarking machine
  learning algorithms, 2017.

\bibitem[Yang(2019)]{yang2019scaling}
Greg Yang.
\newblock Scaling limits of wide neural networks with weight sharing: Gaussian
  process behavior, gradient independence, and neural tangent kernel
  derivation.
\newblock \emph{arXiv preprint arXiv:1902.04760}, 2019.

\bibitem[Zagoruyko and Komodakis(2016)]{Zagoruyko2016WRN}
Sergey Zagoruyko and Nikos Komodakis.
\newblock Wide residual networks.
\newblock In \emph{BMVC}, 2016.

\end{thebibliography}
\newpage
\appendix

\section{Frozen NTK dynamics} \label{app:frozen_ntk}

\begin{figure}[th]
    \centering
    \scalebox{0.8}{\import{figures/}{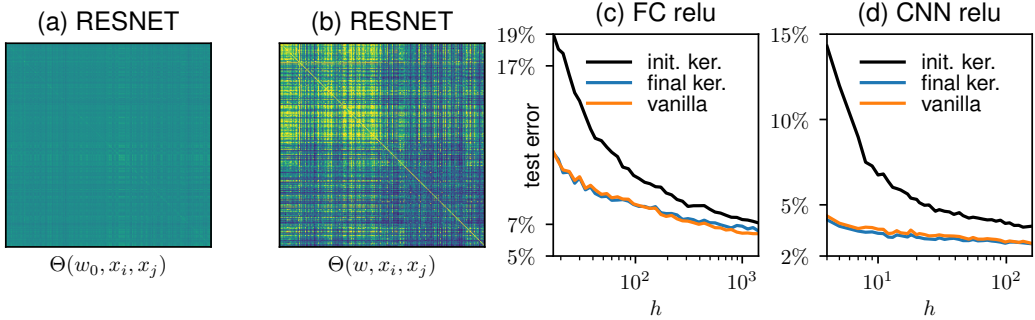}}
    \caption{Gram matrix of the test set at initialization $\Theta(w_0, x_\mu, x_\nu)$ \emph{(a)} and at the end of training $\Theta(w, x_\mu, x_\nu)$ \emph{(b)}, for a wide-resnet 28x10 architecture~\cite{Zagoruyko2016WRN} ($L=25$ hidden layers) trained on a binary version of CIFAR10. The first half of the indices $\mu = 1\dots\nicefrac{n}{2}$ has label $y=1$ and the other half has label $y=-1$. The kernel inflates during learning in a way that depends on the two classes. See \aref{app:frozen_ntk} for a description of the architecture. \emph{(c-d)} Test error \textit{v.s.} the width $h$ for the regular dynamics, the dynamics with the frozen kernel at initialization and the dynamics with the frozen kernel of the end of training. The training performance is captured by the kernel. \label{fig:kernel_dyn}}
\end{figure}

Let us consider the first order approximation $\tilde f_{w_1}(w, x)$ of a model $f(w, x)$ around $w = w_1$,
\be
\tilde f_{w_1}(w, x) = \nabla_w f(w_1, x) \cdot w.
\ee
For instance, $w_1$ can be the value at initialization or at the end of another dynamics. We can then train this linearized model keeping the gradients fixed:
\be
\dot{\tilde{f}}_{w_1}(w) = \nabla_w f(w_1) \cdot \dot w,
\ee
where $\dot w$ depends on the gradient descent procedure. Using gradient descent,
\be
    \dot{\tilde{f}}_{w_1}(w) = -\nabla_w f(w_1) \cdot \frac1n \sum_{(x,y) \in \mc T} \ell'(\tilde f_{w_1}(w, x), y) \nabla_w f(w_1, x)
\ee
By introducing the kernel we can rewrite the previous equation as
\be \label{eq:frozen_dyn}
    \dot{\tilde{f}}_{w_1}(w) = -\frac1n \sum_{(x,y) \in \mc T} \ell'(\tilde f_{w_1}(w, x), y) \Theta (w_1, x)
\ee
where $\Theta$ is the neural tangent kernel defined in \eref{eq:ntkformula}. We call these equations the \emph{frozen kernel dynamics}.

The results presented in \fref{fig:time_dynamics} state that the network learns a kernel during the training dynamics, and that this learned kernel coincides with the frozen kernel in the lazy-training regime as $\sqrt{h}\alpha\to\infty$. Another way to see that the kernel changes during training is to plot the so-called Gram matrix of the frozen kernel, namely the matrix $(\Theta(w,x_\mu,x_\nu))_{\mu,\nu\in\mr{test\ set}}$: in \fref{fig:kernel_dyn} \emph{(a-b)} we show the Gram matrix of the neural tangent kernel evaluated before and after training, where it is clear that there is an emergent structure that depends on the dataset. 

The architecture used in in \fref{fig:time_dynamics} is a resnet based on \cite{Zagoruyko2016WRN}. We use no batch normalization and initialization is as in our fully-connected networks.
The code describing the architecture is available in the supplementary material.

It is interesting to test if the performance of deep networks after learning is entirely encapsulated in the kernel it has learned. We argue that indeed this is the case, and to make our point, we proceed as follows. At any time $t$ during training, we can compute the instantaneous neural tangent kernel $\Theta(w(t))$ as in \eref{eq:ntkformula}; then, we perform a frozen kernel dynamics using that instantaneous kernel, and we evaluate its performance on the test set. In \fref{fig:kernel_dyn} \emph{(c-d)} we plot the test error of a network with $\alpha=1$ (referred to as ``vanilla'') and compare it to the test error of the frozen kernel, both at initialization ($t=0$) and at the end of training. Quite remarkably $\tilde f_{w(t_2)}$ achieves the full performance of $f$.

\section{Dynamics of the Weights} \label{app:weights_evolution}
\begin{figure}[ht]
    \centering
    \scalebox{0.8}{\import{figures/}{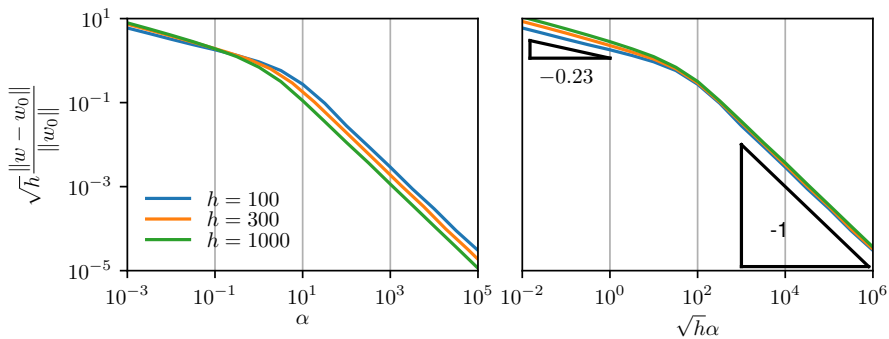}}
    \caption{Relative evolution of the parameters $\sqrt{h} \abs{w-w_0}/\abs{w_0}$ {\it v.s.} $\alpha$ (left) and $\sqrt{h} \alpha$ (right).
    Each measure is averaged over 10 initializations.
    \label{fig:weights_scalings}}
\end{figure}

In \fref{fig:weights_scalings} we plot the rescaled evolution of the parameters $\sqrt{h} \abs{w-w_0} / \abs{w_0}$ versus $\alpha$ \emph{(left)} and versus $\sqrt{h}\alpha$ \emph{(right)} showing that the curves collapse. We find:
\be
\frac{\abs{w-w_0}}{\abs{w_0}} \sim \frac1{h\alpha}
\ee
in the lazy-training regime. It is expected from \eref{aa2}, and the fact that the dynamics lasts ${\cal O}(1)$ in this regime, jointly  implying that the ${\cal O}(h^2)$ internal weights evolve by ${\cal O}(1/(h\alpha))$.
Finally
\be
\label{form}
\frac{\abs{w-w_0}}{\abs{w_0}} \sim \frac{(\sqrt{h}\alpha)^{-b}}{ \sqrt{h}}
\ee
in the feature-training regime, where $b\approx 0.23$ is compatible with $1/(L+1)\approx 0.17$ as proposed in \sref{app:informalargument}. Note that the denominator $\sqrt{h}$ is also expected from \sref{app:informalargument}, where it corresponds to the term $\abs{ w(t_1)-w(0)}$ entering in the definition of $\lambda$. It comes from the fact that $\abs{ w(t_1)-w(0)} \sim \sqrt{h}$, as follows from the ${\cal O}(h^2)$ internal weights evolving by ${\cal O}(1/\sqrt{h})$ on the time scale $t_1$, as can be deduced from \eref{aa2}.

\section{Dynamics of the Output function } \label{app:out}
\begin{figure}[ht]
    \centering
    \scalebox{0.8}{\import{figures/}{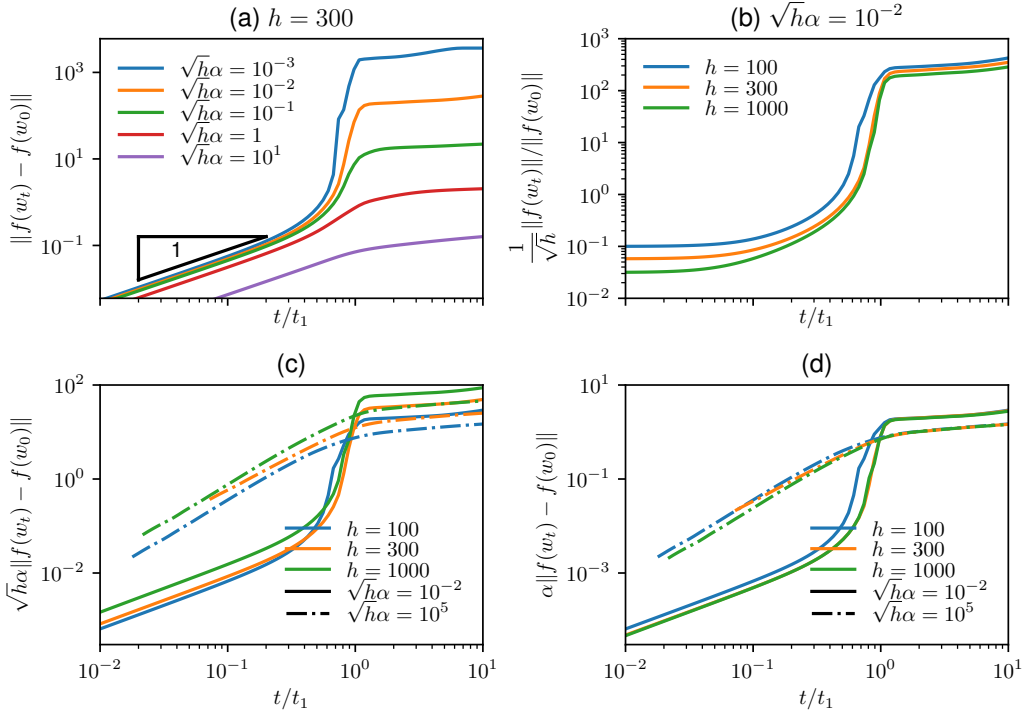}}
    \caption{Different measures of the network norm {\it v.s.} $t / t_1$ for (a) a fixed width $h$ and various $\alpha$, (b,c,d) a fixed value of $\sqrt{h}\alpha$ and various $h$.
    Here $t_1$ is the time at which the loss reduced by half.
    The network used here has $L=2$ hidden layers and uses a \textit{Softplus} activation function. 
    Each curve is averaged along the y axis for 10 realizations.
    \label{fig:dfnorm}}
\end{figure}

To measure the amplitude of the output of a network $f$ we define its norm as follow
\be
\abs{ f(w)} = \sqrt{\langle f(w, x_\mu)^2 \rangle_{\mu \in \mr{test}}}
\ee
In \sref{app:informalargument} we argued that  the dynamics is linear for $t\ll t_1\sim \alpha\sqrt{h}$. \fref{fig:dfnorm} (a,c,d) confirms that the dynamics, characterized by $\abs{ f(w_t) - f(w_0) }$, is indeed linear on a time scale of order $t_1$, independently of the value of $\alpha$ as shown in \fref{fig:dfnorm} (a) or $h$ as shown in \fref{fig:dfnorm} (c,d).

Another important result of \sref{app:informalargument} is that at the end of the linear regime, the output has increased by a relative amount $\sim \sqrt{h}$. This result is confirmed in \fref{fig:dfnorm}(b) showing that $\frac{1}{\sqrt{h}}\abs{ f(w_t)}$ is independent of $h$ for $t\sim t_1$.

These two facts taken together imply:
\be
    \abs{ f(w_t) - f(w_0)} \sim \frac{t}{t_1} \sqrt{h}, \quad t\ll t_1 \text{ in feature training.} \label{eq:dfnorm}
\ee

This prediction is confirmed in \fref{fig:dfnorm} (d) showing  $\alpha \abs{ f(w_t) - f(w_0) } \sim t/t_1 (\sqrt{h} \alpha)$ which must behave as $t/t_1$ if  $(\sqrt{h} \alpha)$ is hold fixed, as is the case in this figure.

\section{Preactivation evolution} \label{app:dot_z}
\begin{figure}[ht]
    \centering
    \scalebox{0.8}{\import{figures/}{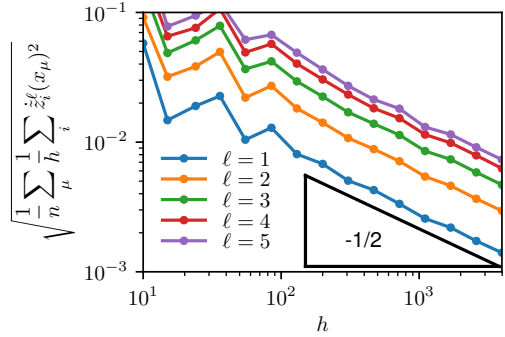}}
    \caption{$\dot{\tilde{z}}$ at $t=0$ {\it v.s.} $h$ for different layers. Each measure is averaged over 5 networks.
    The network used here has $L=5$ hidden layers and uses a \textit{Softplus} activation function. 
    \label{fig:dot_z}}
\end{figure}

\fref{fig:dot_z} shows the amplitude of $\dot{\tilde{z}}$ at initialization.
We measured it using finite-difference method by applying a single gradient descent step.

\section{Shallow network}
\label{app:shallow}
In order to verify the depth dependence of our heuristic predictions about the powerlaw in $\alpha$ of the quantities $\abs{ \Theta - \Theta_0 }$ and $\abs{ w - w_0 }$, we reran the experiment with 2 hidden layers ($L=2$) with the fully-connected network and \textit{Softplus}.
In \taref{tab:thw} we summarize the exponent found numerically, they are compatible the our predictions.

\begin{table}[ht]
    \centering
    \begin{tabular}{c|c|c|c}
        Observable & $L=5$ & $L=2$ & Prediction \\
        $\abs{ \Theta - \Theta_0 }$ & $1.7\ (1.66)$ & $1.25\ (1.33)$ & $\frac{2}{1 + 1/L}$ \\
        $\abs{ w - w_0 }$ & $0.23\ (0.166)$ & $0.35\ (0.333)$ & $\frac{1}{1+L}$ \\
    \end{tabular}
    \caption{Powerlaw dependence in $\alpha$, measure and prediction (in parenthesis) of the exponent $a$ where $O \sim \alpha^{-a}$ for $\alpha \ll 1$}
    \label{tab:thw}
\end{table}

\section{ReLU activation function} \label{app:ReLU}

\begin{figure}[ht]
    \centering
    \scalebox{0.8}{\import{figures/}{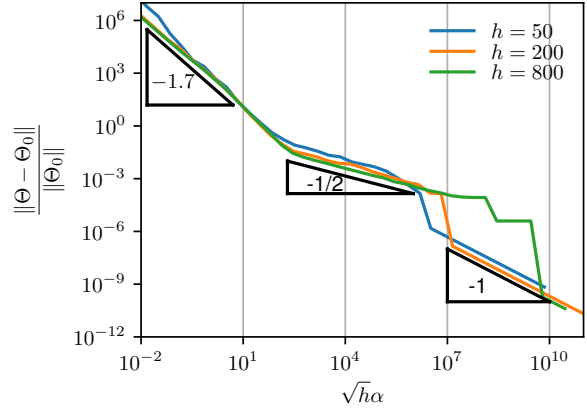}}
    \caption{$\frac{\abs{ \Theta - \Theta_0 }}{\abs{ \Theta_0}}$ {\it v.s.} $\sqrt{h}\alpha$ for different heights. Each measure is averaged over 3 networks.
    The network used here has $L=5$ hidden layers and uses a \textit{ReLU} activation function.
    \label{fig:ReLU_th}}
\end{figure}

\fref{fig:ReLU_th} shows the evolution of the kernel as a function of $\sqrt{h}\alpha$ for a network with \textit{ReLU} activation function.
Differently from the \textit{Softplus} case (see \fref{fig:time_dynamics} (a)), here we observe the existence of three regimes, each characterized by a different power law.
The intermediate regime with slope $-1/2$ is not present for \textit{Softplus}, and it is compatible with the following explanation.
The \textit{ReLU} function $x \mapsto \max(0, x)$ is non differentiable in $x=0$.
It implies that $w \mapsto f(w)$ is not differentiable.
For a finite dataset, $w \mapsto \{ f(w, x_\mu) \}_\mu$ is differentiable only on small patches.
As we can see in \fref{fig:ReLU_th} for large enough $\alpha$, $w$ evolution is so small that $f$ remains in a differentiable patch and we get the predicted result of slope $-1$.
But for intermediate values of $\alpha$, the network change patches a number of times proportional to $\alpha^{-1}$ and assuming that each changes in the kernel induced by these changes of patch are not correlated, their sum scales with $\alpha^{-\nicefrac{1}{2}}$.

\section{Gradient flow verification}
\label{app:sanity_check}

\begin{figure}[ht]
    \centering
    \scalebox{0.8}{\import{figures/}{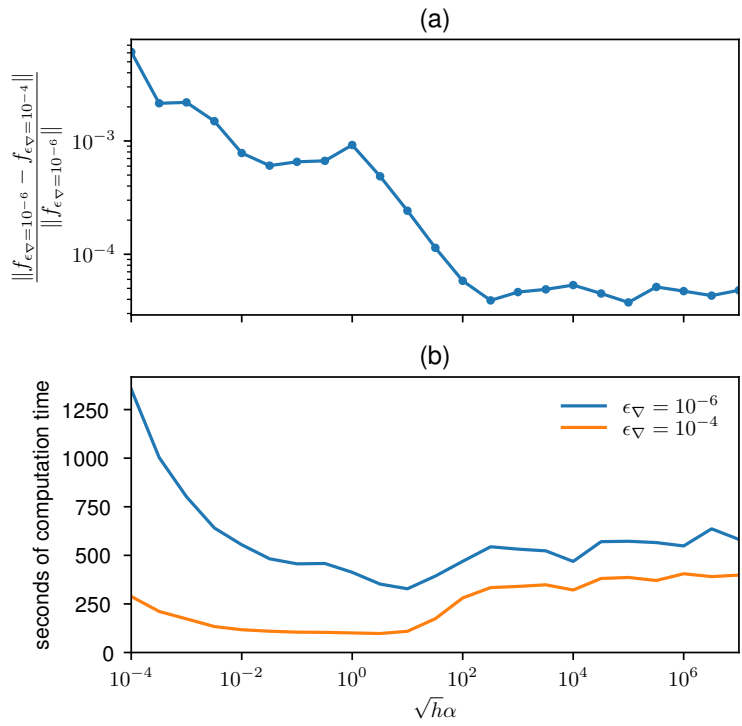}}
    \caption{\emph{(a)} Relative difference of the output for two different values of $\epsilon_\nabla$ \textit{v.s.} $\sqrt{h} \alpha$. \emph{(b)} Relative difference of the output for two different momentum $\tau$ \textit{v.s.} $\sqrt{h} \alpha$. \emph{(c)} Computation time \textit{v.s.} $\sqrt{h} \alpha$, for two different values of $\epsilon_\nabla$. \emph{(d)} Computation time \textit{v.s.} $\sqrt{h} \alpha$, for two different momentum $\tau$.
    \label{fig:sanity_check}}
\end{figure}

As we can see in \fref{fig:sanity_check}, the constraint we put on the relative difference of gradients for the dynamics ensure that the relative difference of the output is smaller than $10^{-3}$ when the constraint is divided by 100. Also we see that the time of convergence is roughly doubled when the constraint is divided by 100.

\section{Variance and the size of the dataset in the lazy-training regime}
\label{app:var_p}

\begin{figure}[ht]
    \centering
    \scalebox{0.8}{\import{figures/}{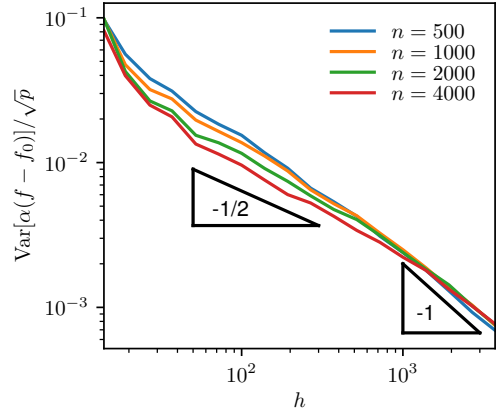}}
    \caption{\label{fig:var_p} Variance of the output in the lazy-training regime ($\sqrt{h} \alpha = 10^6$) \textit{v.s.} the network's width $h$ for different size of trainset $n$. We believe that eventually all the curves asymptote with a slope -1. This asymptote is reached earlier for smaller $n$.}
\end{figure}

In \fref{fig:var_p} the variance is shown as a function of the width, for different sizes of dataset $n$. 
(i) We see the variance reaching the asymptotic behavior of $h^{-1}$ when $n$ is small enough.
(ii) The data also suggest that the variance grows with the size of the trainset like $\sqrt{n}$.

\be
\mr{Var}f(w,x) = \left\langle \left[f(w_i,x_\mu) - \bar{f}(x_\mu)\right]^2 \right\rangle_{\substack{ \mu\in\mr{test} \\ i\in\mr{ensemble} }}\!\!\!.
\ee

\section{$f - f_0$ versus $f$}
\label{sec:alphaf}
\begin{figure}[ht]
    \centering
    \scalebox{0.8}{\import{figures/}{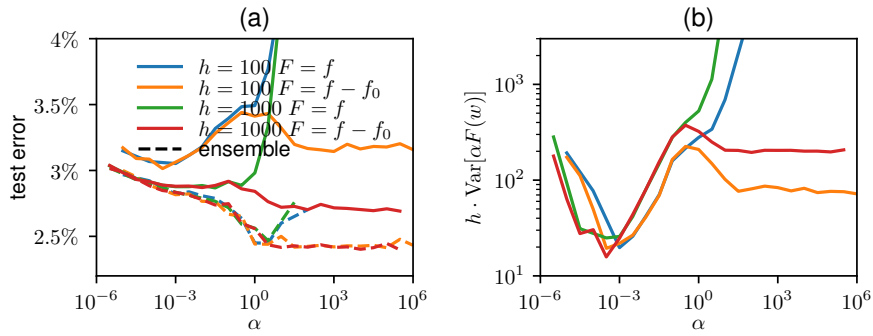}}
    \caption{\label{fig:alphaf}Comparison between the models $\alpha F(w,x) = \alpha(f(w) - f(w_0))$ and $\alpha F(w,x)=\alpha f(w,x)$. The setup is described in \sref{sec:setup}. \emph{(a)} Test error (with ensemble average in dashed line) \textit{v.s.} $\alpha$ for a network of two widths and for the two models. (averaged over 10 initializations) \emph{(b)} The network's width time the variance of the output \textit{v.s.} $\alpha$. (averaged over 10 initializations).
    In the feature regime, both models behave similarly because $f(w_0)$ is negligible compare to $f(w)$.}
\end{figure}

\fref{fig:alphaf} shows the difference between the model $F(w,x) = \alpha f(w,x)$ and $F(w,x) = \alpha (f(w,x) - f(w_0,x))$. Notice that removing the value of the output function at initialization drastically improves the performance of the network for large values of $\alpha$. The generalization error is the same for small $\alpha$.

\section{Effect of biases}
\label{sec:bias}

\begin{figure}[ht]
    \centering
    \scalebox{0.8}{\import{figures/}{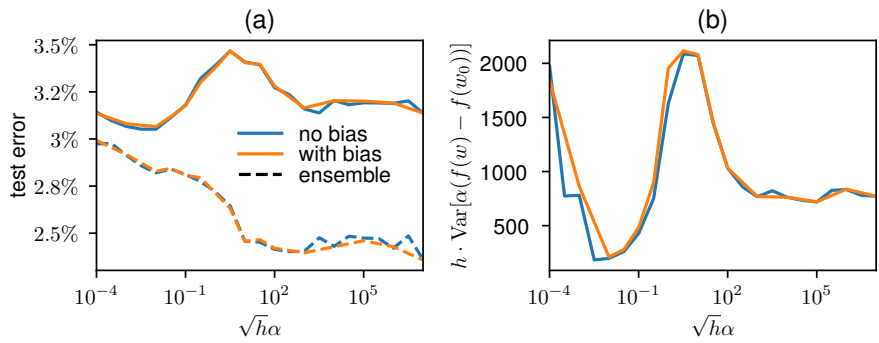}}
    \caption{\label{fig:bias}Same as \fref{fig:vary_alpha} but with biases. \emph{(a)} Test error (single shot and ensemble average) \textit{v.s.} $\sqrt{h} \alpha$ for a network of width $h=100$, averages are over 20 initializations. \emph{(d)} We plot the width times the variance of the output function \textit{v.s.} $\sqrt{h} \alpha$, averaged over 10 initializations.}
\end{figure}

In \fref{fig:bias} we test the helpfulness of introducing biases in hidden layers, by comparing the test error and the variance of the output function in networks that have or have no biases. It turns out that biases are negligible in the present setting. The setup is described in \sref{sec:setup}. A possible reason to use biases would be that ReLU networks without biases are homogeneous functions. This mean that two data $x_\mu, x_\nu$ that are \emph{aligned}, in the sense that $x_\mu = |\lambda| x_\nu$, cannot have different labels. This problem can often be neglected for two reasons: first, in practice the datasets are normalized, so that two points are aligned only if they are identical; second, the pictures in some datasets typically have constant pixels. For instance, in MNIST the top-left pixel in every picture is black. Constant pixels behave effectively as a bias in any hidden neuron.

\end{document}